# Network Level Evaluation of Hang-up Susceptibility of HRGCs using Deep Learning and Sensing Techniques: A Goal Towards Safer Future

Kaustav Chatterjee[1], Joshua Li[*2], Kundan Parajulee[3], Jared Schwennesen[4]

[1]Graduate Research Associate, School of Civil Engineering, Oklahoma State University, Stillwater, 74078, United States of America. Email: kaustav.chatterjee10@okstate.edu

[2] Professor/Williams Professorship, School of Civil Engineering, Oklahoma State University, Stillwater, 74078, United States of America. Email: qiang.li@okstate.edu

[3] Graduate Research Associate, School of Civil Engineering, Oklahoma State University, Stillwater, 74078, United States of America. Email: kundan.parajulee@okstate.edu

[4] Multimodal Division Manager, Multimodal Division, Oklahoma Department of Transportation, Oklahoma City, 73105, United States of America.

## ABSTRACT

Steep-profiled Highway Railway Grade Crossings (HRGCs) pose safety hazards to vehicles with low ground clearance, which may become stranded on the tracks, creating risks of train–vehicle collisions. This research develops a framework for network-level evaluation of hang-up susceptibility of HRGCs. Profile data from different crossings in Oklahoma were collected using both a walking profiler and the Pave3D8K Laser Imaging System. A hybrid deep learning model, combining Long Short-Term Memory (LSTM) and Transformer architectures, was developed to reconstruct accurate HRGC profiles from Pave3D8K Laser Imaging System data. Vehicle dimension data from around 350 specialty vehicles were collected at various locations across Oklahoma to enable up-to-date statistical design dimensions. Hang-up susceptibility was analyzed using three vehicle dimension scenarios: (a) median dimension (median wheelbase and ground clearance), (b) 75-25 percentile dimension (75 percentile wheelbase, 25 percentile ground clearance), and (c) worst case dimension (maximum wheelbase and minimum ground clearance). Results indicate 70, 80, and 95 crossings at the highest hang-up risk levels under these scenarios, respectively. An ArcGIS database and a software interface were developed to support transportation agencies in mitigating crossing hazards. This framework advances safety evaluation by integrating next-generation sensing, deep learning, and infrastructure datasets into practical decision-support tools.

## 1. Introduction

Humped Highway Railway Grade Crossings are the intersections where the profile of the highway adjacent to the railway track can cause different road and safety hazards, some of which include increased risk of train-vehicle collision, vehicle instability while maneuvering over crossing, traffic congestion, and limited accessibility for all vehicle types. Among the different implications of steep profiled crossings, the primary implication is the collision between a vehicle and a train, as vehicles with low ground-clearance or long overhangs or long wheelbases can get stranded on the crossings. A vehicle-train collision can cause damage to properties, derailments, injuries, fatalities, and may release hazardous substances into the environment in the event of a collision between a train and a tanker carrying hazardous material. Table 1 reports some fatal crashes due to the hangup of low-ground clearance vehicles at HRGC and their disastrous consequences on the infrastructure, economy, and the lives of people. For example, the crash due to hangup near Intercession city in Florida in 1993 resulted in property damage of over $14 million and injuries to 59 people.

**Table 1.** Some crashes due to hangup at HRGCs

| Year | Location | Injuries/ Fatalities | Property damages | Literature |
|---|---|---|---|---|
| 1983 | Rowland, North Carolina | 29 injuries | $623,399 | NTSB. (1984) |
| 1983 | Citra, Florida | 59 injuries | $200,119 | Lingenfelter et al. (2018) |
| 1986 | Gary, Indiana | 32 | $110,000 | Lingenfelter et al. (2018) |
| 1987 | Canby, Oregon | 1 | $49,022 | Lingenfelter et al. (2018) |
| 1987 | Seffner, Florida | 17 | $366,349 | Lingenfelter et al. (2018) |
| 1993 | Near Intercession City, Florida | 59 | Over $14 million | NTSB. (1995) |
| 1995 | Graysville, Georgia | 1 | $1,000,000 | Lingenfelter et al. (2018) |
| 1995 | Sycamore, South Carolina | 33 | Over $1 million | NTSB. (2001) |
| 1995 | Milford, Connecticut | 24 | $500,000 | Lingenfelter et al. (2018) |
| 1997 | Jacksonville, Florida | No information | Over $1.4 million | Lingenfelter et al. (2018) |
| 2000 | Glendale, California | No information | Over $2 million | Lingenfelter et al. (2018) |

| 2000 | Intercession City, Florida | No information | $200,000 | Lingenfelter et al. (2018) |
|---|---|---|---|---|
| 2017 | Biloxi, Mississippi | 4 fatalities, 39 injuries | No information | Lingenfelter et al. (2018) |
| 2021 | Thackerville, Oklahoma | 4 injuries | Damage to car-hauler trailer | ABC. (2021) |

Apart from the risk of collision, the steep profiled crossings can disrupt the smooth flow of traffic, causing discomfort and instability for vehicles, especially for vehicles with low ground clearance. The steep profile increases the likelihood of loss of control, accidents, and vehicle damage. Moreover, vehicles stuck on HRGC can lead to congested traffic, causing delays and impacting the efficiency of transportation networks. This can cause frustration for road users and potential risks of secondary accidents or unsafe driving behaviors. Humped crossings also pose challenges to specific vehicles, like low-ground clearance vehicles, motorcycles, bicycles, or those with limited mobility. Ensuring the safety and accessibility of these crossings is crucial for promoting inclusive transportation and accommodating a wide range of vehicles. There are some guidelines for designing the vertical alignment of HRGC that reduces the possibility of hang-up of low clearance vehicles.

The popular design guidelines are from American Association of State Highway and Transportation Official's (AASTHO) *A Policy on Geometric Design of Highway and Streets* (AASTHO. 2018), the U.S. Department of Transportation's Railroad-Highway Grade Crossing Handbook (Ogden and Cooper. 2020), and the American Railway Engineering and Maintenance-of-Way Association's (AREMA) *Manual for Railway Engineering* (AREMA. 2018). Figure 1 shows the vertical alignment of HRGC as per the three different standards. These guidelines are generally followed for the design of a new crossing. However, these guidelines are sometimes not followed during the construction and maintenance activities of the railway track, resulting in the formation of a hump crossing.

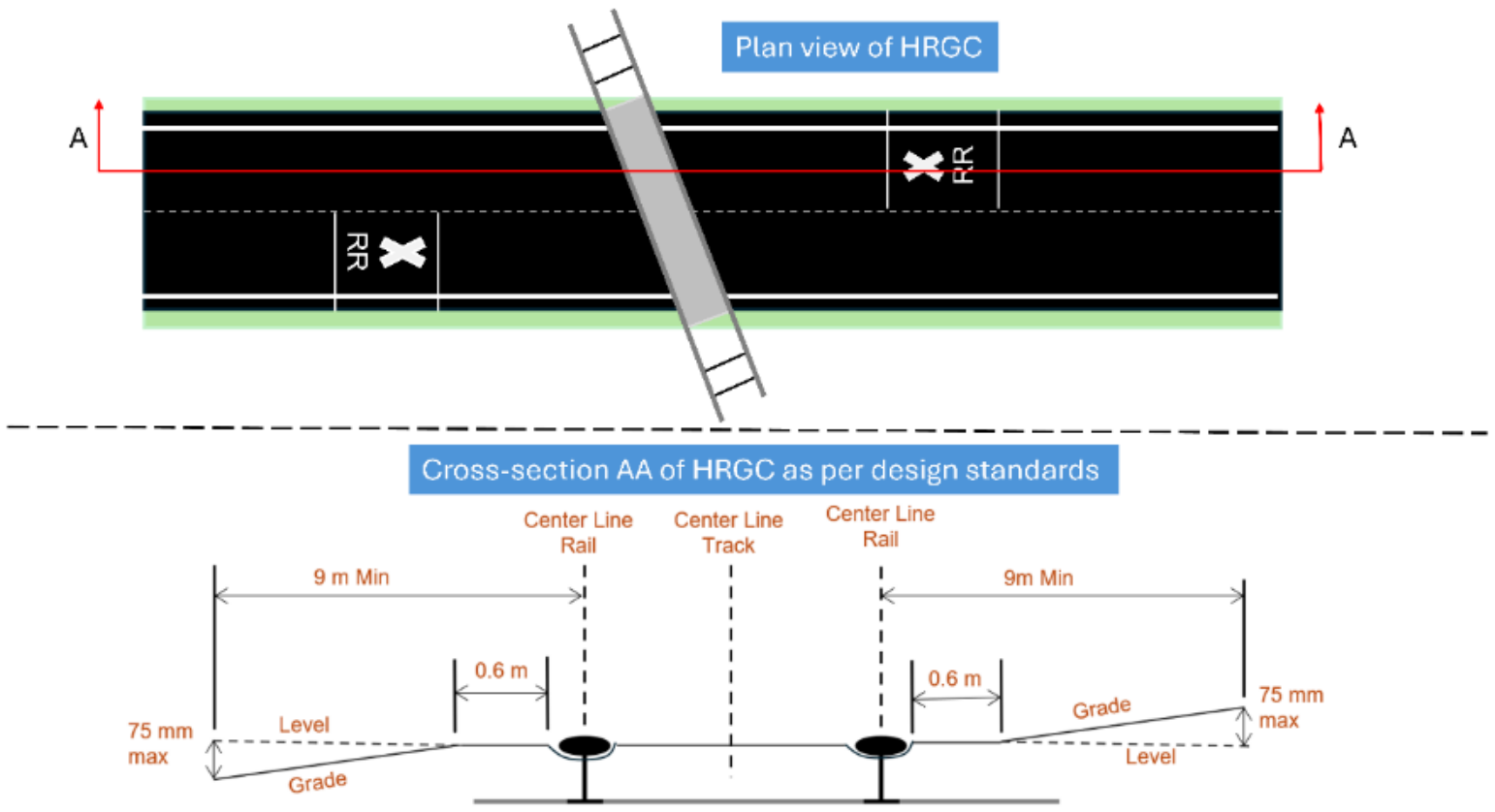

**Figure 1.** Design Guidelines for vertical alignment of HRGCs (AASTHO. 2018, AREMA. 2018, Ogden and Cooper. 2018)

Different researchers (Eck and Kang. (1991, 1992)) highlighted the issue of hang-up of low-clearance vehicles and developed formulated approaches to mitigate it. For example, researchers developed and employed different techniques like physical models (Mutabazi and Russell. 2003), Light Detection and Ranging (LiDAR) (Khattak et al. 2015, Lee and Khattak. 2016, Liu et al. 2017), and Unmanned Aerial Vehicles (UAV) (Chatterjee et al. 2025) for measuring the profile of hump crossings. Moreover, other researchers like Clawson. (2002), and French et al. (2003) worked on developing the design vehicle of low ground clearance vehicles.

Although the above-mentioned research works indicated great promise at the time, several limitations affect their applicability in current practice. For instance, the physical models developed by Mutabazi and Russell (2003) were constrained due to their heavy weight and requirements for on-site traffic control during data acquisition, thereby limiting their scalability and feasibility. In addition, LiDAR-based (Khattak et al. 2015, Lee and Khattak. 2016, Liu et al. 2017) data acquisition approaches provide high-resolution data but are limited by their high expense for widespread implementation. The shortcoming associated with UAV-based profile acquisition is that the UAV data were affected by the presence of trees surrounding the HRGC and vehicles on the HRGC. Given these limitations, there is a requirement for methods that can measure the profile of HRGCs using advanced sensing technologies and deep learning techniques to provide a more accurate and comprehensive assessment of hang-up susceptibility at the HRGCs. These gaps motivate the use of Pave3D8K Laser Imaging at traffic speeds, coupled with hybrid deep learning, to provide a cost-effective and scalable evaluation method.

The novelty of this study lies in establishing a framework for network-level evaluation of hangup susceptibility of HRGCs by integrating IMU-GPS sensor data and hybrid deep learning models. Moreover, this is the first study to build an Oklahoma-specific design vehicle dimension dataset for performing comprehensive analysis of hangup susceptibility with different levels of risks. This study also developed an interactive software interface facilitating different transportation organizations like Department of Transportation (DOTs) and railroad companies to perform hangup analysis accurately and efficiently. Lastly, this study contributes to safety of infrastructure and people by developing an ArcGIS database containing hangup levels of different crossings. The ArcGIS database would facilitate transportation organizations in selecting crossings for maintenance operations and aid drivers of low-ground clearance vehicles in safer route selection by avoiding high-profiled crossings.

## 2. Methodology

HRGC profile data were collected using the walking profiler SurPRO and Pave3D8K laser Imaging Technology; SurPRO walking profiler provided centimeter-resolution ground truth but required traffic control, Pave3D8K Laser Imaging System enabled high-speed data collection across with IMU–GPS and laser sensors. Profiles from both systems formed paired datasets for deep learning model development. A hybrid LSTM–Transformer deep learning model was developed to reconstruct HRGC profiles from the inertial measurement unit (IMU) and GPS data collected by Pave3D8K at highway speed. Thereafter, low clearance vehicle dimension data were acquired from trailer manufacturers and trailer repairing centers in Oklahoma, and statistical analysis was performed on the vehicle dimension data to obtain the design dimension. Using HRGC profiles from a deep learning model, instrumentation data and design dimensions of vehicles from statistical analysis, hangup analysis of the RedRock Corridor in Oklahoma was performed. Lastly, an ArcGIS database and SAFEXING software interface was developed to facilitate easy accessibility of the research accessible to different transportation organizations.

## 3. Field data collection

### *3.1 Study Area: RedRock Corridor*

The Red Rock Corridor is a key rail section operated by the BNSF Railway Company in Oklahoma that extends from the Texas border to the Kansas border. This corridor has a significant role in supporting the transportation infrastructure and economy of the state. It can primarily be divided into two subdivisions: the southern section, from the Texas border to Oklahoma City, that accommodates both freight and passenger rail services, including Amtrak's Heartland Flyer, which operates two trains a day, and the northern section, from Oklahoma City to the Kansas border, exclusive to freight traffic.

In this research, the Red Rock Corridor was selected as the case study due to its multimodal nature and its significance as one of Oklahoma's most vital railway corridors in terms of economic impact. Since this corridor includes both freight and passenger trains, the operational complexity increases for the railway industry, and therefore it can become a suitable candidate for crossing evaluation.

### *3.2 Ground truth profile data collection: The walking profiler*

The SurPRO, walking profiler is a widely used instrument for measuring the profiles of various transportation infrastructures, including highways, roads, and airfield pavements. It measures accurate profile of HRGCs using an inclinometer at a resolution of 1 cm. The data from this device can be extracted in multiple formats .ERD, .PPF, and .PRO. This device was employed to collect ground truth profile data from HRGCs.

The major limitation of this device is the necessity for traffic control, risks to personnel acquiring profiles on HRGCs, and the slow speed of data acquisition. These shortcomings can be alleviated by deploying Pave3D8K Laser Imaging vehicle for profile data collection. Figure 2 shows the field data collection utilizing the walking profiler and Pave3D8K Laser imaging data vehicle.

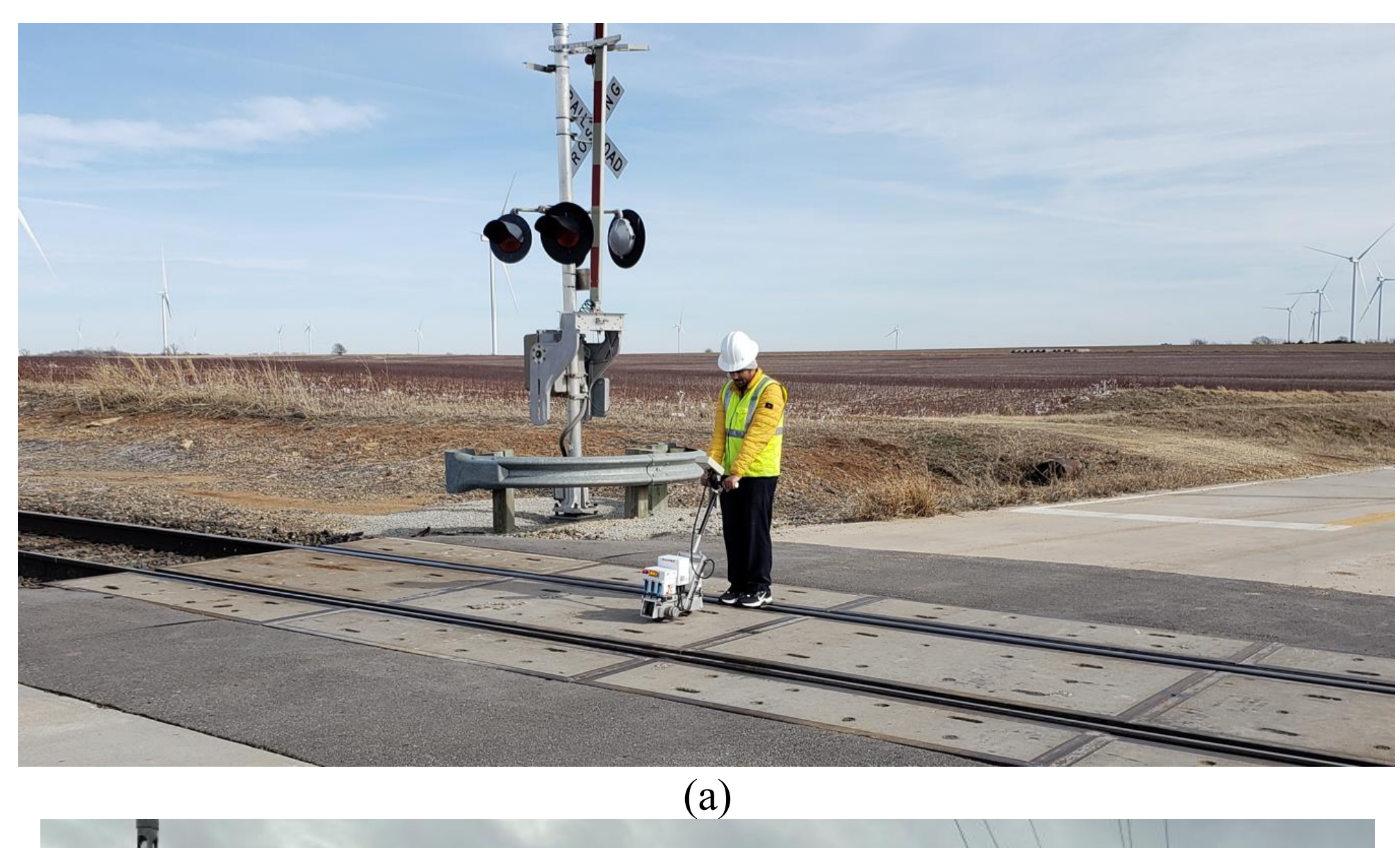

(a)

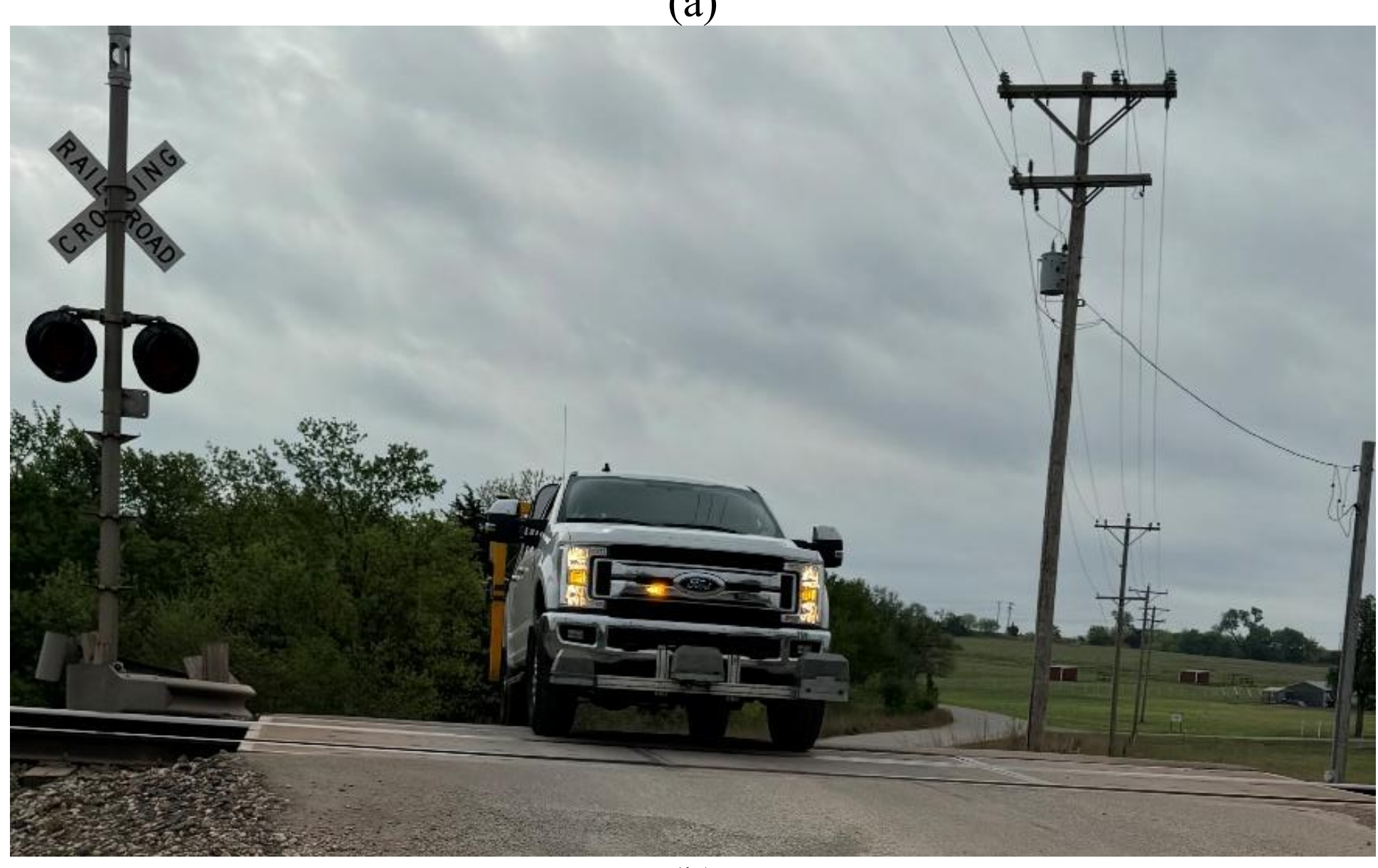

(b)

**Figure 2.** Field data collection of HRGC profiles (a) walking profiler (b) Pave3D8K Laser Imaging Data vehicle

### *3.3 Instrumentation profile data collection: Pave3D8K Laser Imaging Data Vehicle*

Pave3D8K Laser Imaging Data Vehicle (Chatterjee et al. 2024, Parajulee et al. 2025, Chatterjee et al. 2026) consists of arrays of different sensors, including Inertial Measurement Unit (IMU), Global Position System (GPS), 3D camera, laser, and distance measuring instrument, used for acquiring high-resolution pavement surface data at driving speed. In this research, IMU and GPS data were used to estimate the profile of HRGCs.

The IMU-GPS data provides information about the pitch, roll, latitude, longitude, altitude, and acceleration in X, Y, and Z directions. The altitude data from the GPS sensor provides information about the profile of HRGC. However, a significant difference was observed between the profile from the ground-truth profiler and the GPS-measured profile. This problem was fixed by leveraging sequence-to-sequence deep learning models for profile determination.

### *3.4 Dataset for model development*

The dataset considered in this research consisted of sequential time-series data obtained from different HRGCs. The input sequence of the dataset consisted of seven vehicular parameters: accelerations in three directions (X, Y, Z), pitch, roll, vehicle speed, and altitude, collected using an integrated IMU-GPS sensor system. The output sequence was the profile data measured using a high-resolution walking profiler, serving as ground-truth for deep learning model during training and validation. Each HRGC sequence contained approximately 2500 data points, resulting in an input sequence of shape (2500 x 7) from the IMU-GPS sensor and a corresponding output sequence of shape (2500 x 1) from the walking profiler.

### *3.5 Data augmentation for deep learning model development*

Deep learning models are capable of learning highly complex patterns, but their performance depends on the availability of large datasets. To ensure the optimal performance of the deep learning model, the original dataset was augmented using two data augmentation techniques in order to generate training, validation, and test samples. A schematic representation of the augmentation process is shown in figure 3.

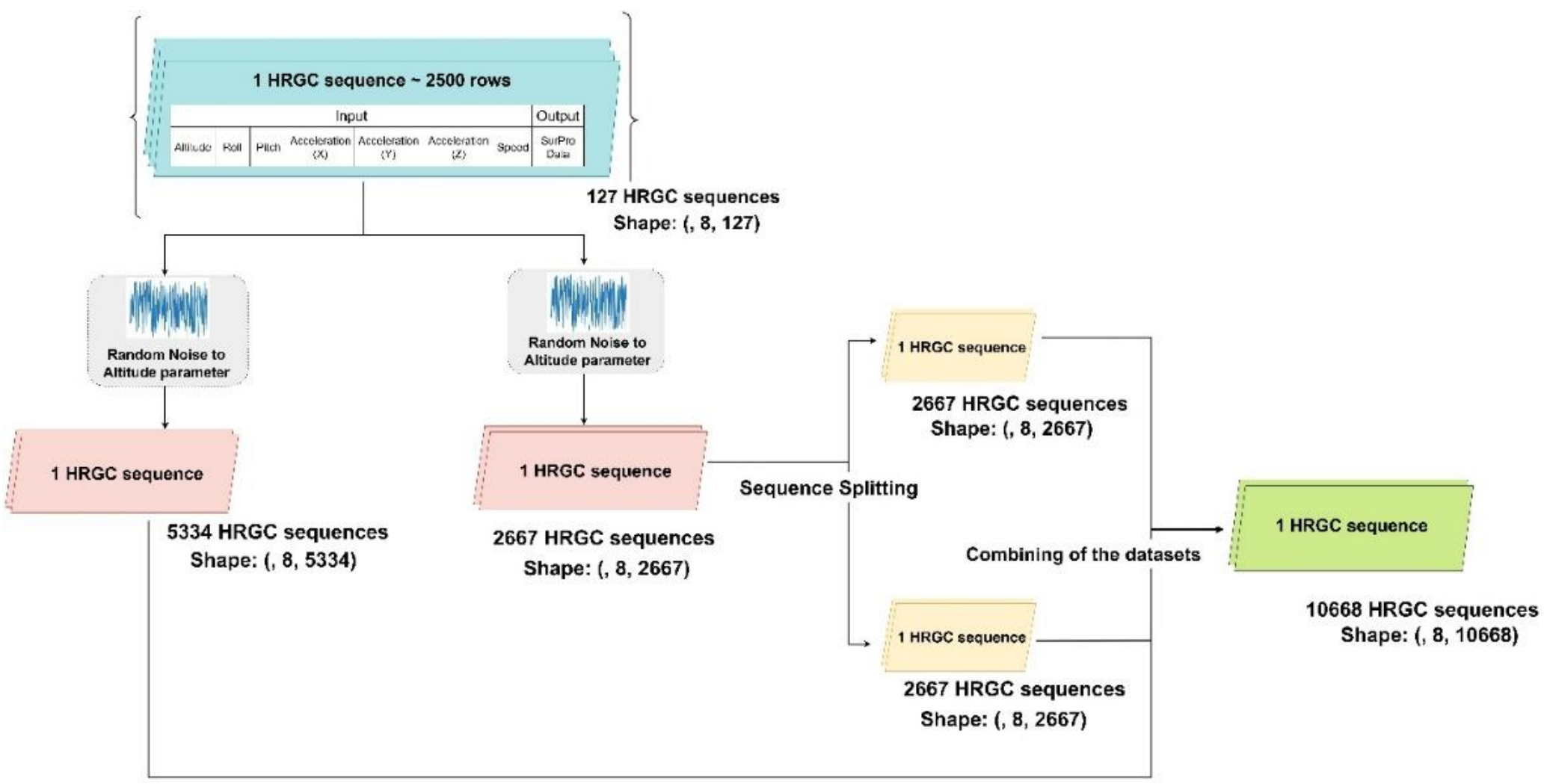

**Figure 3.** Schematic representation of data augmentation techniques

In the first technique, random noise was added to the GPS-based profile to generate 5,334 sequences. This noise was generated using a normal distribution with a mean of zero, and a standard deviation equal to four percent of the range (i.e., the difference between the highest and lowest values) of each sequence. In the second technique, random noise was introduced into the profile data to generate 2667 sequences. Thereafter, each of the 2667 sequences were split into two new sequences: one sequence composed of the odd-indexed points and the other sequence composed of the even-indexed points. This process generated 5334 sequences from the original 127 HRGC sequences. Using both techniques, 10668 data sequences were created for model development; 7888, 1387, and 1393 sequences were used for training, testing, validation, respectively.

## 4. Hybrid Deep Learning Model Development

A special type of deep learning model, sequence-to-sequence models, was used to determine the profile of HRGCs from IMU-GPS data. The sequence-to-sequence model receives the IMU-GPS sequence as input, and the output from the model is the profile of HRGCs. In recent times, sequence-to-sequence deep learning models received widespread adoption in machine translation and time series forecasting. The advancement in sequence-to-sequence modelling was accelerated by the advancement in computer algorithms and computational resources.

In this research, hybrid deep learning models were developed by fusing the transformer and LSTM architecture. The philosophy behind the development of hybrid models is that the model can take advantage of the working principles of both architectures, the LSTM would facilitate in understanding the short-term dependencies in the HRGC data, while the transformer layer would facilitate in understanding long-term dependencies of the data.

### *4.1 LSTM*

LSTM (Hochreiter and Schmidhuber. 1997) is widely used in sequence-to-sequence deep learning modelling, consisting of three different gates: forget gate, input gate, and output gate. The forget gate carries out the job of discarding irrelevant information from the memory cell, the input gate performs the job of controlling the amount of new data to be included in the LSTM memory cell, and the output gate performs the job of controlling the extent of data that will be delivered to the next memory cell. The mathematical operations related to the three gates are presented in equation (1) to equation (8). Figure 4 shows a schematic representation of the LSTM architecture.

$$f_y = \sigma(W_f.\left[t_{y-1}, a_y\right] + b_f) \quad (1)$$

$$\sigma(y) = \frac{1}{1+e^{-y}} \quad (2)$$

The forget gate, sigmoid activation function, weight matrix and bias vector are represented by $f_y$, $\sigma(y)$, $W_f$, and $b_f$, respectively.

$$i_y = \sigma(W_i.\left[t_{y-1}, a_y\right] + b_i) \quad (3)$$

$$C'_y = tanh(W_c.\left[t_{y-1}, a_y\right] + b_c) \quad (4)$$

$$tanh(t) = \frac{e^y - e^{-y}}{e^y + e^{-y}} \quad (5)$$

The input gate and candidate cell state $i_y$ and $C'_y$, respectively. $W_i$, $W_c$ represents the bias vector of the input gate and candidate cell, respectively and $b_i$, $b_c$ represents the input gate and candidate cell, respectively.

$$C_y = f_y \times C_{y-1} + i_y \times C'_y \quad (6)$$

Equation (6) represents the critical cell state update operation. The new cell state $C_y$ is derived by multiplying the previous state $C_{y-1}$ by the forget gate and adding the candidate state scaled by the input gate.

$$O_y = \sigma(W_o.[t_{y-1}, a_y] + b_o) \ (7)$$

$$h_y = O_y \times \tanh(C_y) \ (8)$$

The output gate ($O_y$) and the hidden state ($h_y$) are computed in Equation (7) and Equation (8), respectively. $W_o$ and $b_o$ represent the weights and biases associated with the output gate.

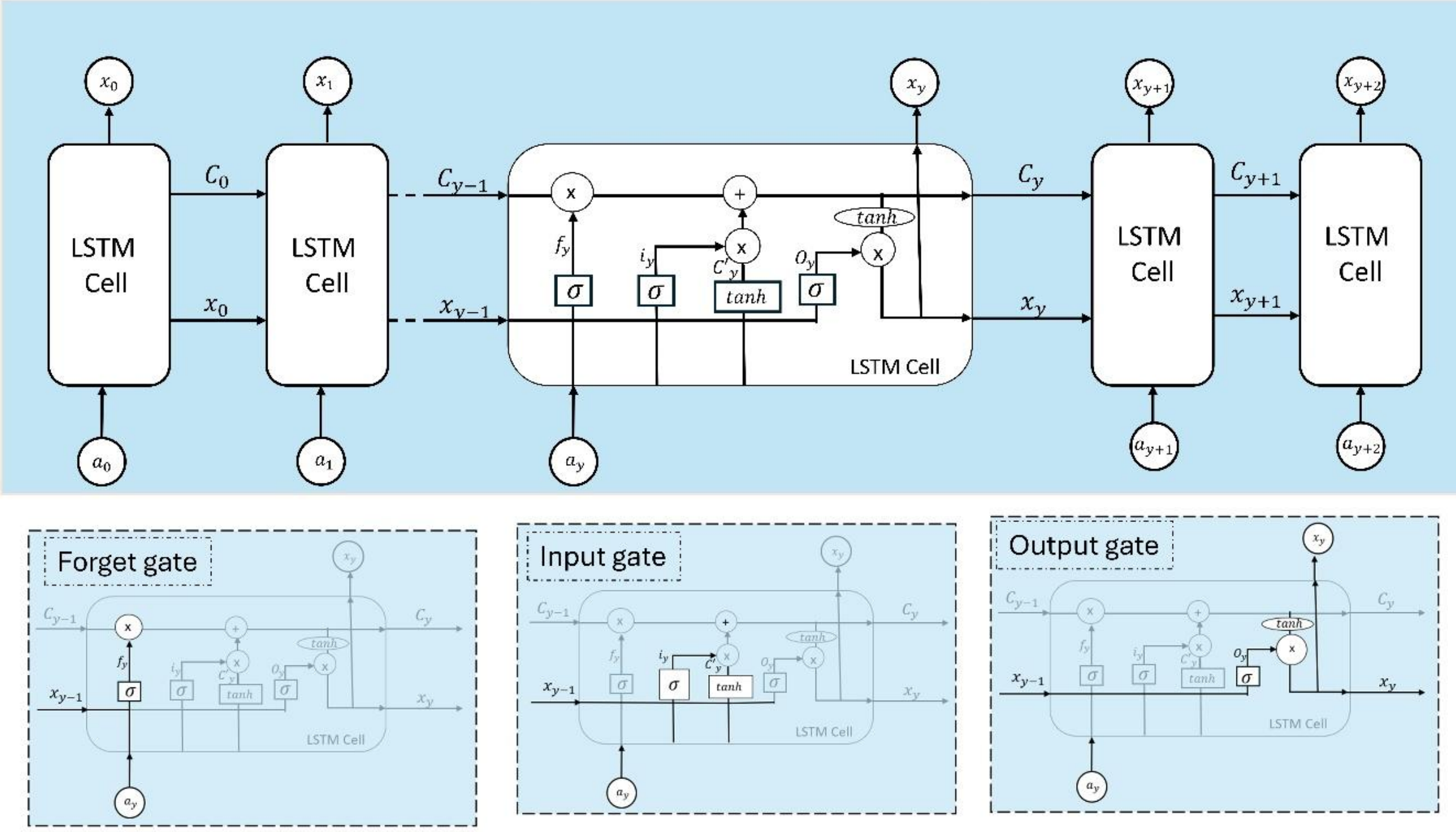

**Figure 4.** Schematic representation of architecture of LSTM network

### *4.2 Transformer*

Transformer architecture was developed by Vaswani et al. (2017) for different machine translation tasks. It is extensively used in sequence-to-sequence learning due to its competence in capturing long-term dependencies of the sequences and improved processing speed by virtue of parallel processing. The original transformer architecture developed by Vaswani et al. (2017) comprised of two main blocks: Encoder and Decoder. In contrast, the transformer architecture designed in this research comprises an encoder block; an analogous approach was employed by Anik et al. (2024) in their study.

Figure 5 shows the architecture of the transformer block, comprising two main components: positional encoding and multi-head attention. The positional encoding block furnishes positional information for different dataset points of input sequences; this is an integral part of the architecture since the transformer architecture is devoid of the recurrent or convolutional operation. The mathematical definition of positional encoding is as follows:

$$PE(position, 2l) = sin\ (position/10000^{2l/d})\ (9)$$

$$PE(position, 2l + 1) = cos\ (position/10000^{2l/d})\ \ (10)$$

Here, $position$ is the token in the input sequence, $l$ is the dimension of the position index vector and $d$ is the dimension of the representation.

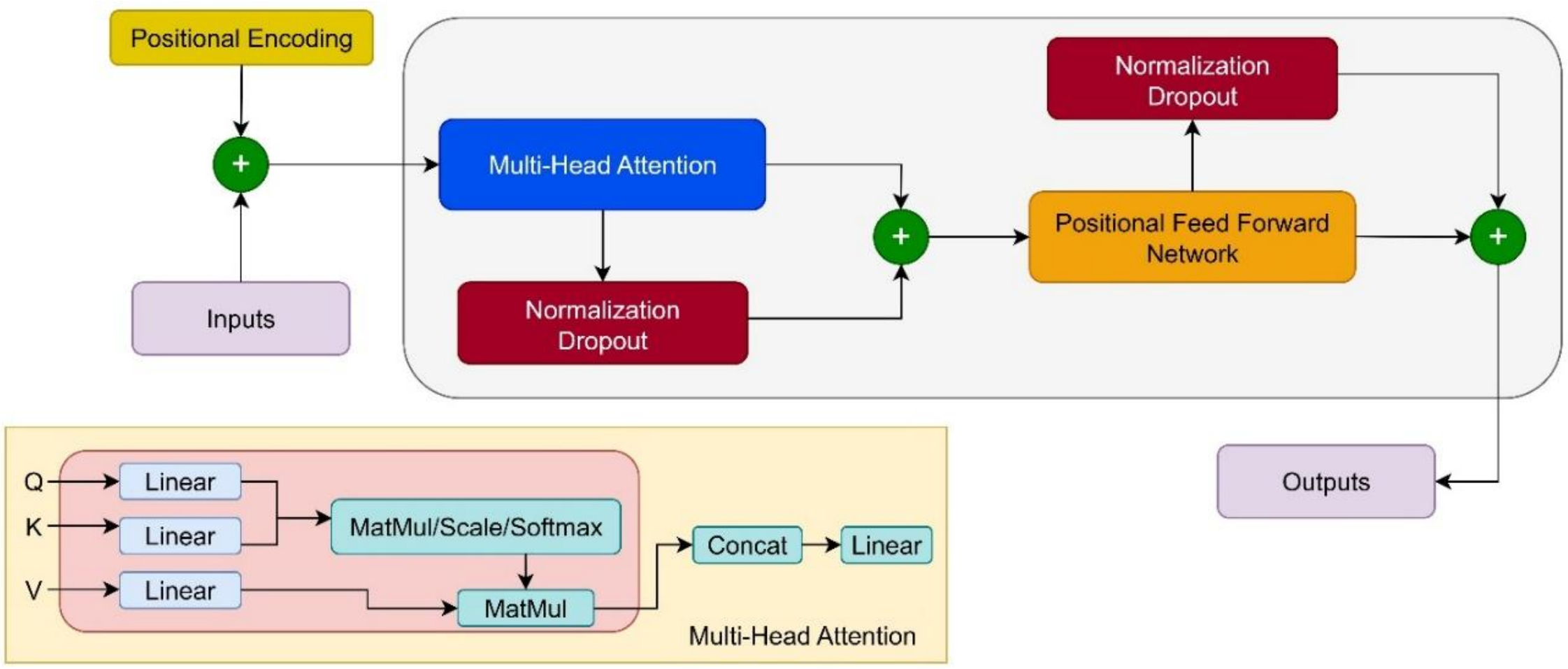

**Figure 5.** Architecture of Transformer network

Multi-head attention facilitates the transformer architecture to comprehend the relationship between different components of the input sequence. The multi-head attention was calculated by combining the output vector from single-head self-attention, and thereafter performing a linear transformation of the combined value. Mathematically, single-head self-attention is defined as follows:

$$Attention(X) = \sigma(\frac{QK^T}{\sqrt{d_k}})V\ \ (11)$$

$$Q = XW_q; K = XW_k; V = XW_v\ (12)$$

Here $X$ is the position encoded input, $\sigma$ represents the Softmax function; $Q, K,$ and $V$ represents the query, key, and value, respectively, $d_k$ represents the dimension of key, and $W_q$, $W_k$, and $W_v$ represents the weight matrices, and T represents the transpose matrices.

### *4.3 Model Architecture and Performance*

The LSTM-transformer hybrid model was developed by placing the LSTM and transformer architecture in parallel, enabling the transformer and LSTM architecture to process data in parallel without interaction, LSTM block can identify the short-term dependencies in the data, whereas transformer block can identify the long-term dependencies in the data. Fusing the output from each architecture enables the model to identify both short-term and long-term dependencies in the HRGC sequence data. Figure 6 shows the architecture of the developed model.

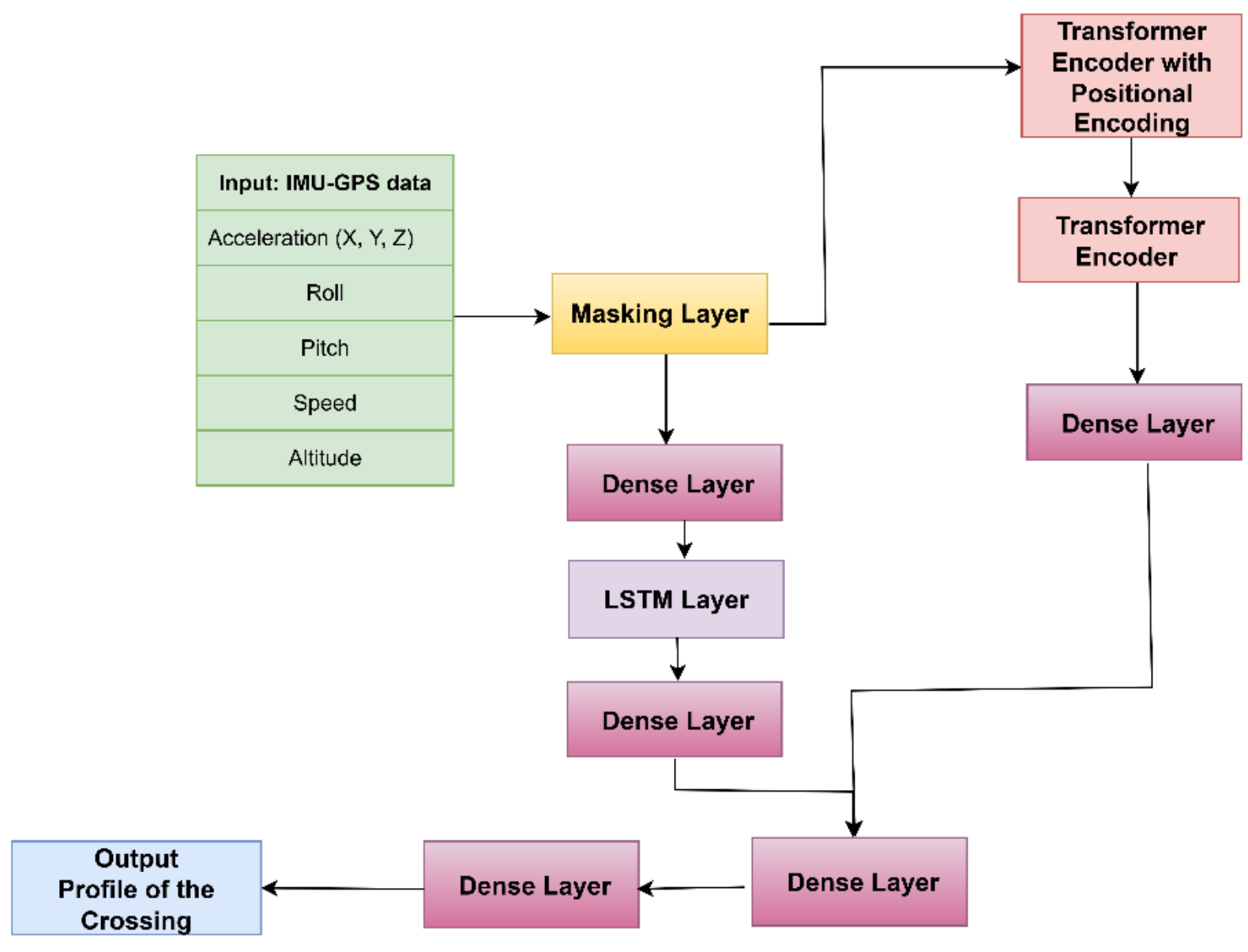

**Figure 6.** Architecture of the hybrid model (Chatterjee et al. 2026)

To quantitatively assess the predictive accuracy of the proposed model, two standard regression metrics were employed: Root Mean Square Error (RMSE) and Mean Absolute Error (MAE). These metrics serve to rigorously quantify the magnitude of deviation between the model's estimates and the ground truth measurements obtained via the walking profiler. The mathematical representations of these performance metrics are presented in Equations 13 and 14:

$$RMSE = \sqrt{\frac{1}{S}\sum_{j=1}^{S}(a_j - \widehat{b_j})^2} \tag{13}$$

$$MAE = \frac{1}{S}\sum_{j=1}^{S}|a_j - \widehat{b_j}| \tag{14}$$

Where $RMSE$ denotes the root mean square error, $MAE$ represents the mean absolute error, $a_j$ corresponds to the values predicated by the model, $\widehat{b_j}$ represents the actual ground truth data, and $S$ indicates the number of observations in the sequence.

Table 2 summarizes the quantitative results of this evaluation. The results indicate that the model achieves satisfactory precision, as evidenced by the small magnitude of both RMSE and MAE metrics. Following this primary assessment on the test dataset, the study further investigated the model's generalizability to verify its robustness against data variability and its stability under varying operational conditions.

**Table 1.** Performance of model on training, validation and test dataset

| Training | | Validation | | Test | |
|---|---|---|---|---|---|
| RMSE (m) | MAE (m) | RMSE (m) | MAE (m) | RMSE (m) | MAE (m) |
| 0.07 | 0.05 | 0.07 | 0.05 | 0.07 | 0.05 |

### *4.4 Generalizability of model results*

For the model's generalizability test, an evaluation was conducted using a distinct set of nine HRGC sequences that were strictly excluded from the training, validation, and testing phases. The quantitative results of this external validation are summarized in Table 3. The model demonstrated high stability, yielding RMSE and MAE values comparable to those achieved during the model development phases, thereby confirming its strong generalizability to unseen data.

Furthermore, the study assessed the models' resilience during low-resolution data scenarios, simulating scenarios involving low-cost instrumentation with minimized acquisition frequencies. This was achieved by down-sampling the nine HRGC sequences by a factor of 2 to emulate low-resolution inputs. As demonstrated in **Table 3**, the framework maintained high predictive performance even with the sparse dataset, showcasing its ability to successfully reconstruct accurate HRGC profiles despite the reduction in input resolution.

**Table 2.** Generalization of model

| Down sampling factor | RMSE (m) | MAE (m) |
|---|---|---|
| No | 0.07 | 0.06 |
| 2 | 0.07 | 0.06 |

## 5. Design Vehicle Dimension

Vehicle dimension is a critical component in assessing the hang-up susceptibility of low ground clearance vehicles at high-profiled HRGCs. The vehicle dimensions considered include the length of the wheelbase, rear overhang, front overhang, and ground clearance at each of these three locations. Figure 7 provides a schematic representation of a truck highlighting these dimensions.

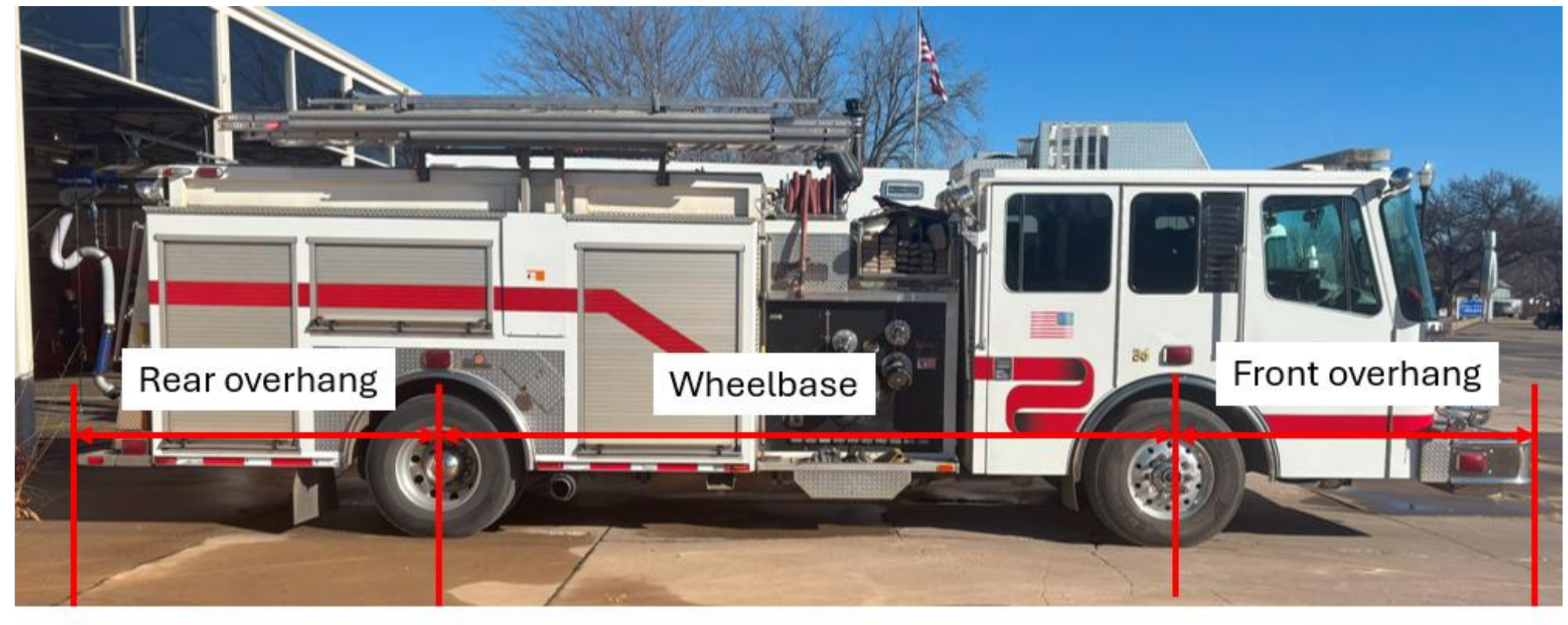

**Figure 7.** Schematic representation of design vehicle dimension

The wheelbase refers to the distance between the axles of the vehicle or trailer, front overhang is defined as the distance from the front end of the vehicle to the center of the forward axle, and the rear overhang is measured from the center of the rear-most axle to the rear end of the vehicle. Ground clearance is the vertical distance between the ground and the bottommost point of the vehicle body, measured at three specific points: at the wheelbase between the axles and under both the front and rear overhang.

Vehicle dimension data are available from different literature (Clawson. 2002, French et al. 2003), however, those data represent vehicle parameters from earlier periods, and the vehicle designs and dimensions may have evolved over time. Design vehicle dimension data can also be obtained from trailer manufacturer specification sheets. In this research, updated information on design vehicle dimensions was obtained from the field measurements at different locations in Oklahoma. This approach was followed to obtain a true representation of the dimensions of trailers travelling in Oklahoma.  Field data collection of design vehicle dimension was also performed by researchers (Clawson. 2002, French et al. 2003).

Dimension data was analyzed using a Python code, and different statistical parameters for each design dimension of every vehicle type were computed. Table 4 lists the statistical parameters related to the wheelbase of low clearance vehicles. The wheelbase length and ground clearance for different class 9 vehicles such as belly dump trailer, drop deck trailer, flatbed trailer, low boy trailer, tanker and class 9 trailer ranges between 6.71 m to 12.5 m and 0.18 m to 0.64m, respectively, while the wheelbase length and ground clearance for different vehicles with class 3, 4, and 5 such firetrucks, recreational vehicles (RVs), school bus, trailer with car or truck ranges between 3.12 m to 11.63 m and 0.15 m to 0.61 m, respectively.

**Table 4.** Design vehicle dimensions at wheelbase

| Vehicle | Sample | Wheelbase (m) | | | Ground clearance (m) | | |
|---|---|---|---|---|---|---|---|
| | | 50% | 75% | max | Min | 25% | 50% |
| Belly Dump | 24 | 10.06 | 10.29 | 11.13 | 0.23 | 0.25 | 0.32 |
| Drop Deck | 28 | 9.75 | 10.36 | 11.28 | 0.25 | 0.3 | 0.41 |
| Firetruck | 8 | 5.41 | 5.54 | 5.97 | 0.27 | 0.32 | 0.34 |
| Flatbed | 53 | 10.97 | 10.97 | 11.89 | 0.25 | 0.38 | 0.43 |
| Low Boy | 10 | 10.36 | 10.78 | 11.89 | 0.18 | 0.21 | 0.23 |
| Recreational Vehicle | 50 | 5.89 | 6.79 | 7.87 | 0.15 | 0.23 | 0.3 |
| Rear Dump | 27 | 8.53 | 8.84 | 9.45 | 0.36 | 0.41 | 0.46 |
| School Bus | 46 | 7.01 | 7.01 | 7.16 | 0.15 | 0.23 | 0.23 |
| Tanker | 22 | 9.3 | 10.61 | 11.58 | 0.28 | 0.37 | 0.41 |
| Class 9 box trailer | 43 | 10.97 | 11.58 | 12.5 | 0.2 | 0.32 | 0.38 |
| Car/truck with trailer | 23 | 6.05 | 8.27 | 11.63 | 0.25 | 0.28 | 0.3 |
| Class 5 Truck | 4 | 6.91 | 7.01 | 7.32 | 0.33 | 0.37 | 0.38 |

Among the different low-ground clearance vehicles considered in this study, front overhang was observed in fire trucks and school buses, both averaging 2.25 m; however, fire trucks had a greater variability of 0.30 m compared to the school buses (0.12 m). Rear overhangs, on the other hand, were observed in a broader range of vehicles, ranging between 1.52 m and 3.71 m for the class 9 vehicles and between 1.27 m and 5.97 m for the class 3, 4, and 5 vehicles. The ground clearance at the rear overhang for class 9 vehicles and class 3, 4, and 5 vehicles ranged between 0.18 and 0.56 m and 0.17 m to 0.66 m, respectively.

## 6. Hangup analysis of RedRock Corridor

Hangup analysis was performed for the RedRock Corridor in Oklahoma across different HRGCs with four levels of risk categorized from Level 1 (lowest risk) through Level 4 (highest risk), Table 5 lists the criteria for different hangup levels along with the hangup analysis results. Three types of analyses were conducted with varying vehicle dimensions: (a) median vehicle dimension, (b) 75-25 percentile analysis, and (c) worst case analysis. Analysis with median dimension was performed to simulate the hangup susceptibility of crossing with a reliability of 50 percent, 75-25 percentile analysis was performed using 75 percentile of wheelbase dimension and 25 percentile of ground clearance dimension to obtain higher reliability than the median analysis, and worst-case analysis was performed to simulate worst case situation using maximum wheelbase and minimum ground clearance. Different transportation organizations should adopt results from worst case analysis in route selection in the event of a low-ground clearance carrying sensitive chemicals or hazardous materials.

The median analysis indicates that the number of crossings in hangup levels 1, 2, 3, and 4 is 6, 13, 19, and 70, respectively, indicating 70 crossings are dangerous for low-ground clearance vehicles and may result in vehicle-train crashes. The number of level 4 crossings obtained from 75-25 percentile analysis and worst-case analysis are 80 and 95, respectively, indicating a higher reliability than the median analysis. Oklahoma DOT can use the information obtained from the hangup analysis to flag different crossings as susceptible to hangup and post warning signs before those crossings. The warning signs would caution drivers of low-ground clearance vehicles from traversing across those crossings.

**Table 5.** Results of hang-up analysis

| Level of hangup | Criteria | Analysis | | |
|---|---|---|---|---|
| | | Median | 75-25 percentile | Worst case |
| 1 | δ => 0.1016 m | 6 | 3 | 0 |
| 2 | 0.1016 m > δ => 0.0508 m | 13 | 10 | 3 |
| 3 | 0.0508 m > δ => 0 inches 0.00 m | 19 | 15 | 10 |
| 4 | δ < 0.00 m | 70 | 80 | 95 |

## 7. ArcGIS Database

The results obtained from the median hangup analysis were used to develop an ArcGIS database, consisting of the level of hangup of different vehicles for a particular crossing. Different transportation organizations such as Department of Transportation (DOTs) can use this ArcGIS database as an efficient, fast, and cost-effective technique of obtaining information of hangup level of different crossings without encountering the challenges faced in performing hangup analysis. Additionally, DOT can use the database to visualize the locations of crossings with different hangup levels and selecting crossings for maintenance action.

DOTs can extract information on level of hangup from the database and include them on maps and circulate the maps to different trailer companies. The trailer companies can employ these maps in safer route selection by avoiding high-profile crossings. Figure 8 shows a screenshot from the developed ArcGIS database for four different design vehicles, with green, yellow, orange, and red dots representing hangup levels 1, 2, 3 and 4, respectively.  Low-boy trailers are susceptible to hangups at 64.81% of the crossings, while the fire trucks can safely maneuver across 80.5% of the crossings, as indicated by the green dots.

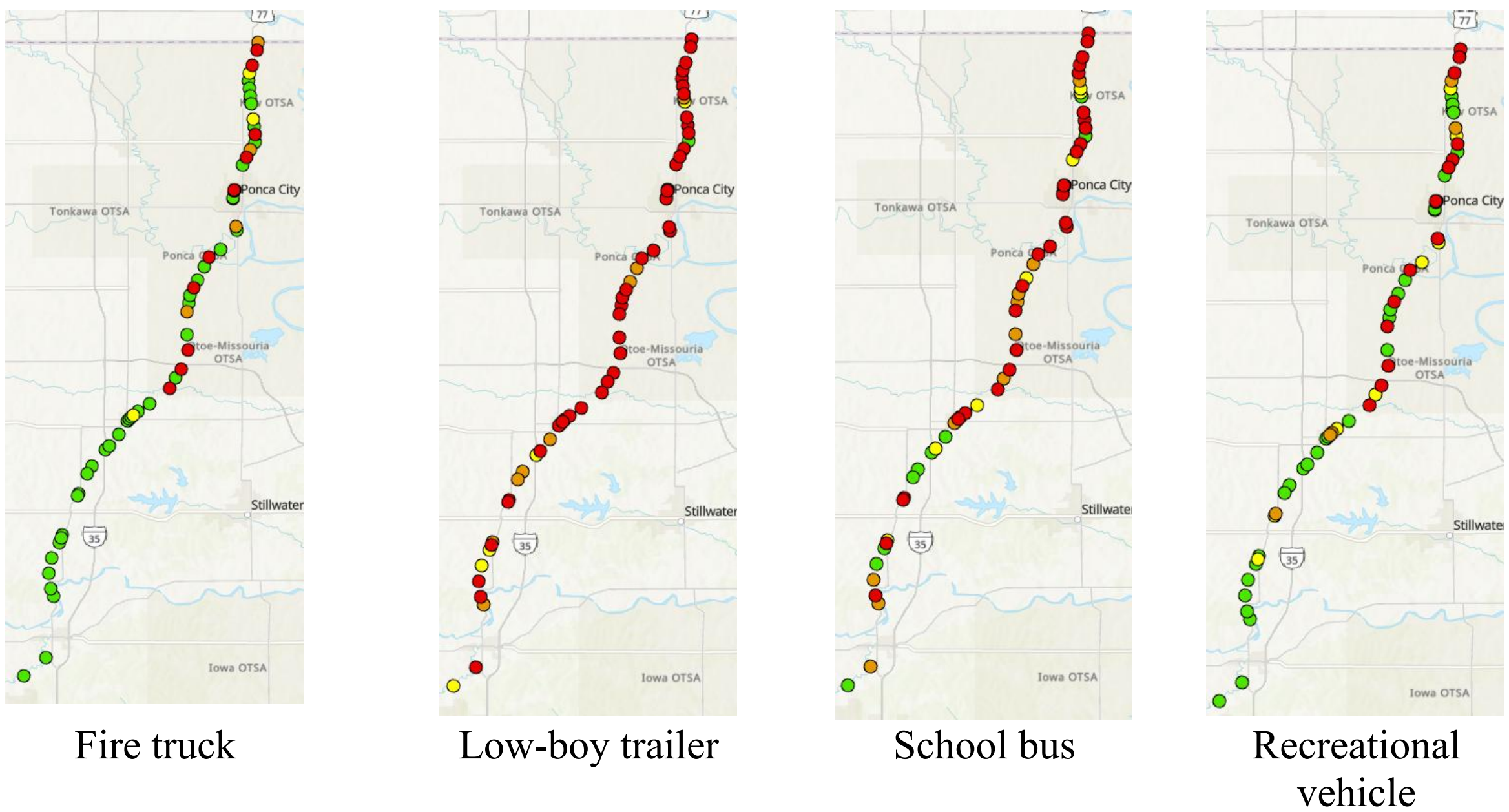

Fire truck | Low-boy trailer | School bus | Recreational vehicle

**Figure 8.** Visualization of hangup level using ArcGIS database

## 8. Software for Hangup Analysis: SAFEXING

The software (SAFEXING) was developed to facilitate easy accessibility of this research to different transportation organizations, such as the Department of Transportation. The software consists of five different tabs: Introduction, Crossing Details, Profile Visualization, Low Clearance Vehicles, and Hang-up Analysis. Figure 9 shows the screenshot of the Crossing Details and Hangup Analysis tab of the software. The first tab, "Introduction," is the home page of the software, providing background information on the hang-up analysis, including an introduction to the hang-up analysis problem and design guidelines for vertical alignments of HRGCs.

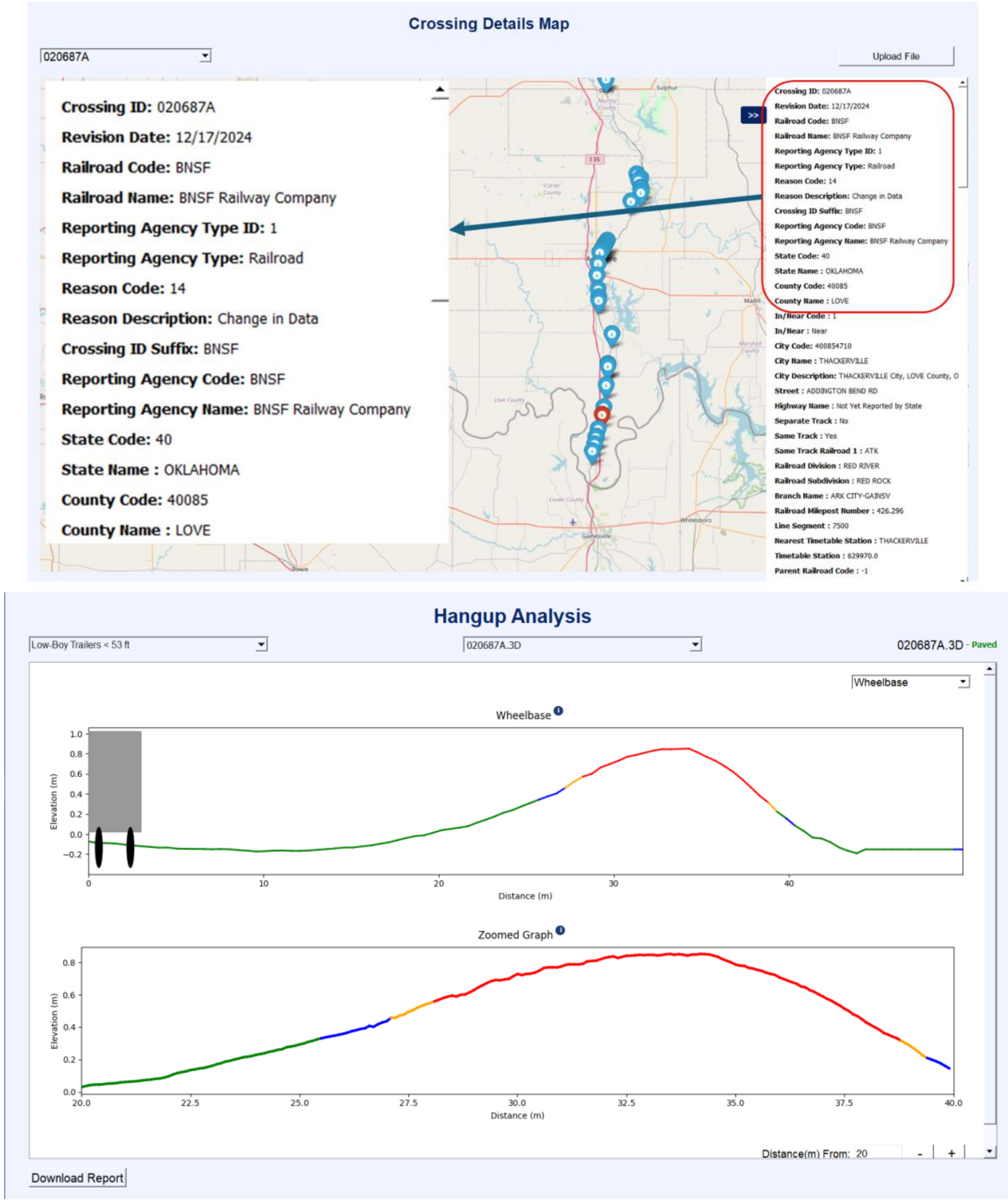

**Figure 9.** Screenshot of the software (a) crossing details, (b) hangup analysis

Crossing Details tab presents information such as crossing ID, county code, city code, street, highway name, railroad division, or railroad subdivision, about the different crossings in the Red Rock Corridor and their locations on an interactive map.

The Low Clearance Vehicles tab presents details of the dimensions of different types of vehicles. The hangup analysis tab enables the user to perform the hangup analysis of hump crossing, the input of the software is the data from the IMU-GPS sensor, and the output of the software is the level of hang-up at each point on the profile.

### *8.1 Application of SAFEXING*

To demonstrate the benefits and contribution of this software towards the safety of people and infrastructure, a case study analysis was conducted on one crossing (crossing id: 020687A). This specific crossing was selected in reference to a historical incident where a car-hauler trailer collied with an Amtrak train at this crossing in October 2021 resulting in the destruction of the trailer and hospitalization of five passengers with minor injuries[5].

The hangup analysis of the crossing was performed using the profile obtained from the deep learning model and design dimensions of Low-boy trailer. The results of the hangup analysis are presented in figure 9b, suggesting the crossing to be a hump crossing. Various scratch marks on the crossing surface visually confirmed the hump of the crossing. Since the crossing is hang-up susceptible, warning signs should be posted to alert drivers of low ground clearance vehicles.

## 9. Conclusions

This is the first study to establish a framework for network-level evaluation of hangup susceptibility of HRGCs by incorporating instrumentation data from Pave3D8K Laser Imaging Technology and hybrid deep learning models. The edge of this framework over conventional infrastructure-monitoring techniques are no requirement of traffic regulations during profile data collection of HRGCs, quick and accurate evaluation of hang-up susceptibility, user-friendly software interface for hangup analysis, comprehensive assessment of risk of hangup using different vehicle dimensions, and easy visualization and extraction of hang-up levels information using ArcGIS database.

This research commenced with HRGC profile data acquisition from different crossings in Oklahoma using walking profiler, served as the ground truth data, and using Pave3D8K Laser Imaging Technology, instrumentation data collected at traffic speed. Subsequently, hybrid deep learning model using LSTM and Transformer architecture was developed to determine profile of HRGC using Pave3D8K Laser Imaging Technology data. The model development facilitated the profile data in a cost-effective and efficient manner without the requirement of traffic control.

Vehicle dimension data was collected from different locations in Oklahoma and was subsequently analyzed using a Python code. Three sets of design vehicle dimensions were developed: median of wheelbase and ground clearance, 75 percentiles of wheelbase and 25 percentiles of ground clearance, and maximum wheelbase and minimum ground clearance. The median vehicle dimension was developed considering a risk of 50%, the 75-25 percentile vehicle dimension was developed to reduce the risk as compared to the median analysis, and worst-case vehicle dimension was developed to reduce the risk to a low level. Developed design vehicle dimensions of low-ground clearance facilitate in performing hangup analysis with different degree of reliability.

Hangup analysis of RedRock Corridor was performed using the profile obtained from the deep learning model, Pave3D8K Laser Imaging Technology, and design vehicle dimension data from statistical analysis. The number of hangup susceptible crossings from the median, 75-25 percentile analysis and worst-case analysis are 70, 80 and 95, respectively. The results of the hangup analysis will facilitate the Oklahoma DOT in highlighting the hangup susceptible crossing and posting appropriate warning signs in front of those crossings. The warning signs would alert drivers of low-ground clearance about the impending risks from those crossings.

An ArcGIS database was developed to assist transportation agencies in visualizing the risks of hangup and facilitating crossing selection for maintenance operation without encountering the challenges associated in hangup analysis. Moreover, DOTs can disseminate the information from the ArcGIS database to different trailer organizations, facilitating easier route selection with reduce risk of hangup.

A software was developed to facilitate easy accessibility of the developed framework to different transportation organizations. It can be used to obtain different information about HRGC, visualize the profile of HRGC, information on the latest design dimensions of low ground clearance vehicles, and perform hangup analysis. Overall, the framework developed in this study will facilitate saving lives of people, reducing injuries and hospitalization, and reduce risks to highway and railway infrastructure.

**Acknowledgement**

The author(s) disclosed receipt of the following financial support for the research, authorship, and/or publication of this article. The authors would like to acknowledge the financial support received from Oklahoma Department of Transportation for this research work. The authors are thankful to Rohith Dacharla for their contribution in this research.

**Authors contribution statement**

Study conception and design: KC, JL; Field data collection: KC, JL, KP; Data analysis: KC, JL; Manuscript preparation: KC, JL, KP, JS. All authors reviewed the manuscript.

**Data Availability**

The data that support the findings of this study are available from the corresponding author, Joshua Li, upon reasonable request.

**Funding Declaration**

This research was funded by the Oklahoma Department of Transportation through the State Planning and Research (SP&R) Program, Project No. 2296, titled "*Highway and Rail Intersection Hump or High-Profile Crossing Problems*"